\documentclass[10pt,twocolumn,letterpaper]{article}

\usepackage{iccv}              
\usepackage{times}
\usepackage{epsfig}
\usepackage{graphicx}
\usepackage{amsmath}

\usepackage{booktabs}
\usepackage{amssymb}
\usepackage{bbding}
\usepackage{pifont}
\usepackage{wasysym}
\usepackage{utfsym}
\usepackage{fontawesome}

\usepackage{makecell}

\usepackage{silence}
\usepackage{color}

\usepackage{multirow}

\newcommand{\figref}[1]{Fig.~\ref{#1}}
\newcommand{\tabref}[1]{Tab.~\ref{#1}}

\newcommand{\equref}[1]{Eqn.~(\ref{#1})}

\newcommand{\myPara}[1]{\noindent\textbf{#1}}
\newcommand{\sArt}{state-of-the-art~}

\def\eg{\emph{e.g.,~}}
\def\ie{\emph{i.e.,~}}

\newcommand{\ourMthd}{TAR3D}
\newcommand{\nameofposE}{TriPE}

\definecolor{iccvblue}{rgb}{0.21,0.49,0.74}
\usepackage[pagebackref,breaklinks,colorlinks,allcolors=iccvblue]{hyperref}

\def\paperID{1123} 
\def\confName{ICCV}
\def\confYear{2025}

\title{\ourMthd{}: Creating High-Quality 3D Assets via Next-Part Prediction}

\author{
Xuying Zhang$^{1}$\textsuperscript{*}\textsuperscript{\S},  
Yutong Liu$^3$\textsuperscript{*},  
Yangguang Li$^{4,5}$,  
Renrui Zhang$^4$,  
Yufei Liu$^6$, 
Kai Wang$^{1}$, \\
Wanli Ouyang$^{4,6}$, 
Zhiwei Xiong$^3$, 
Peng Gao$^6$, 
Qibin Hou$^{1,2}$\textsuperscript{\dag}, 
Ming-Ming Cheng$^{1,2}$\\ 
\small $^1$VCIP, CS, Nankai University
\quad $^2$NKIARI, Shenzhen Futian 
\quad $^3$USTC \\
\small $^4$CUHK MMLab 
\quad $^5$VAST 
\quad $^6$Shanghai AI Lab\\
{\tt\small Homepage:~\url{https://github.com/HVision-NKU/TAR3D}}
}

\begin{document}
\twocolumn[{%
    \renewcommand\twocolumn[1][]{#1}%
    \maketitle
    \vspace{-33pt}
    \begin{center}
      \captionsetup{type=figure}   
      \footnotesize
       \includegraphics[width=0.93\linewidth]{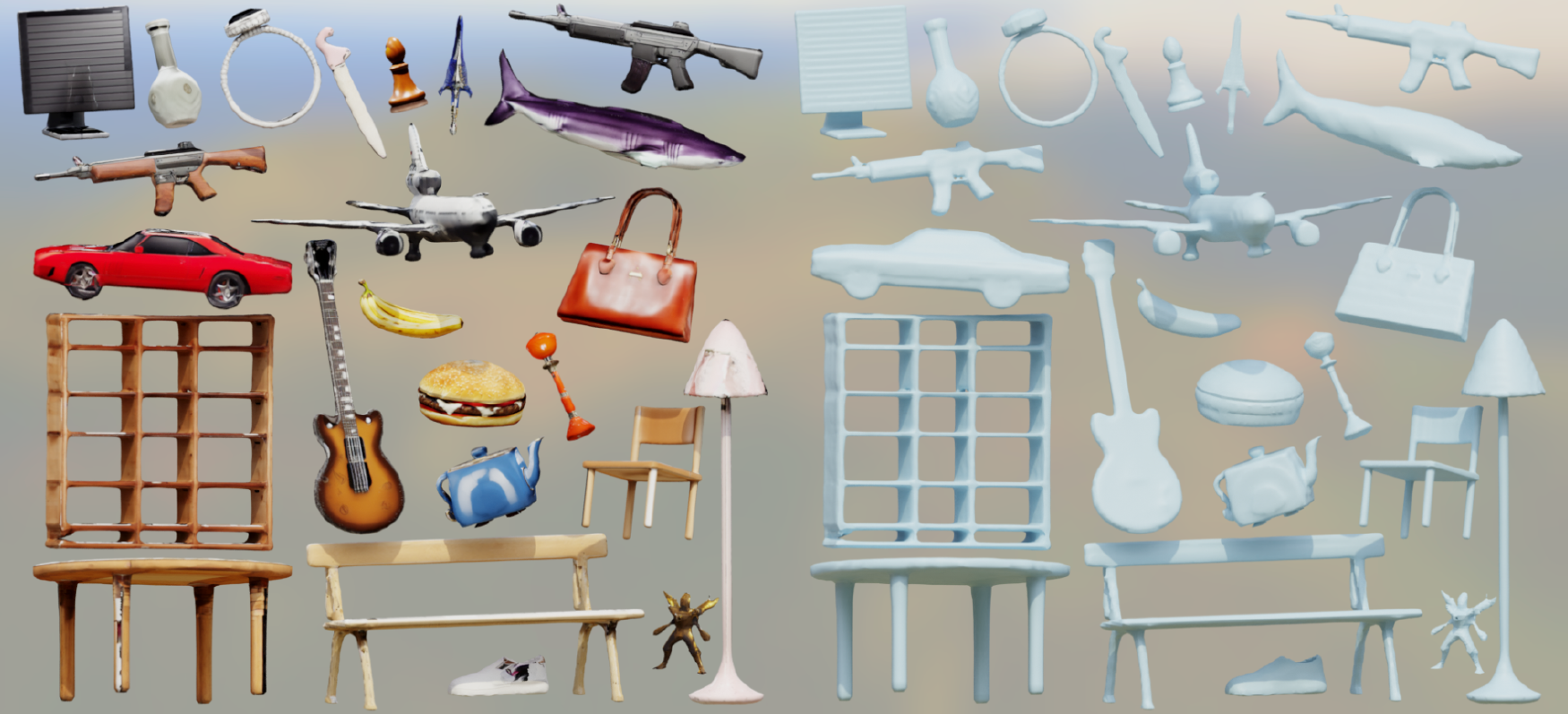}
       \vspace{-5pt}    
      \captionof{figure}{
        3D asset gallery generated by the proposed \ourMthd{} model.
        We employ SyncMVD~\cite{liu2024text} as the texture synthesizer.
      }
      \label{fig:intro}
    \end{center}%
}]  

\renewcommand{\thefootnote}{\fnsymbol{footnote}}
\footnotetext[0]{\textsuperscript{*}Equal contribution. ~~~~~~~~
    \textsuperscript{\dag}Corresponding author.
}
\footnotetext[0]{
    \textsuperscript{\S}Work done during the internship at Shanghai AI Lab. 
}


\begin{abstract}

\vspace{-7pt}
%
We present \ourMthd{}, a novel framework that consists of a 3D-aware Vector Quantized-Variational  AutoEncoder (VQ-VAE) and a Generative Pre-trained Transformer (GPT) to generate high-quality 3D assets.
The core insight of this work is to migrate the multimodal unification and promising learning capabilities of the next-token prediction paradigm to conditional 3D object generation.
To achieve this, the 3D VQ-VAE first encodes a wide range of 3D shapes into a compact triplane latent space and utilizes a set of discrete representations from a trainable codebook to reconstruct fine-grained geometries under the supervision of query point occupancy.
Then, the 3D GPT, equipped with a custom triplane position embedding called \nameofposE{}, predicts the codebook index sequence with prefilling prompt tokens in an autoregressive manner so that the composition of 3D geometries can be modeled part by part.
Extensive experiments on several public 3D datasets demonstrate that \ourMthd{} can achieve superior generation quality over existing methods in text-to-3D and image-to-3D tasks.
%

\end{abstract}

\vspace{-15pt}
\section{Introduction}
\label{sec:intro}
\vspace{-2pt}

Conditional 3D object generation aims to generate high-quality 3D assets that semantically conform to the given prompts, \eg 2D images, and text.
It has witnessed significant progress with substantial advances in diffusion-based methods, including Score Distillation Sampling (SDS) optimization~\cite{poole2022dreamfusion,chen2023fantasia3d}, multiview synthesis~\cite{liu2023zero,shi2023zero123++,xu2024instantmesh}, and 3D-aware diffusion generation~\cite{zhang2024clay,wu2024direct3d,lan2025ln3diff}.
However, establishing a unified model across various modalities poses huge challenges due to the different paradigms between diffusion models and large language models (LLMs)~\cite{achiam2023gpt,bai2023qwen,chen2024internvl,luo2024mono}.

More recently, pioneering works like MeshGPT~\cite{siddiqui2024meshgpt} and MeshXL~\cite{chen2024meshxl} attempt to introduce the autoregressive manner of LLM into 3D mesh generation.
However, they struggle in a sequence length of 9 times the face number, which is attributed to their quantization of the vertices of mesh faces.
Such an excessively long sequence length limits their applicability in industrial 3D assets with hundreds of thousands of faces.
This motivates us to investigate whether it is possible to encode a mesh with an arbitrary number of faces into a fixed-length sequence to alleviate the computational burden of autoregressive modeling.
%

%
%
%

To address this question, the key is to efficiently encode the 3D representations such that the sequence length is not proportional to the number of faces.
Triplane representation provides such a good property.
Different from other 3D representations like unordered point clouds or verbose voxels, this representation is able to balance storage efficiency with strong expressiveness by compressing 3D information into three 2D feature maps with a fixed size.
Due to the high-level 2D compositions, it fits discretization better and can also take advantage of the strategies from image quantization methods, e.g., LLamagen~\cite{sun2024autoregressive}.
Based on triplane representation, we propose \ourMthd{}, a novel 3D autoregressive framework consisting of a 3D VQ-VAE and a 3D GPT.

The 3D VQ-VAE encodes the 3D shapes into compact triplane features, from which a trainable codebook of context-rich 3D geometry parts is employed to acquire a set of discrete embeddings.
In this way, the 3D meshes can be represented as a feature sequence with triplane-size length, regardless of the number of faces, reducing the reliance on huge GPU resources.
The quantized latent representations are then decoded to a neural occupancy field for 3D reconstruction.
To achieve information exchange among different planes, we propose incorporating feature deformation and attention mechanism designs during the decoding process for fine-grained geometry details.
The 3D GPT is driven by prefilling prompt embeddings to model the codebook index sequence corresponding to quantized triplane features, enabling conditional 3D object generation in an autoregressive manner.
To preserve more spatial information, during generation, we also custom-craft a 3D positional encoding strategy dubbed \nameofposE{}, in which the 2D positional information of each plane and the 1D ones between the same position of the three planes are organically fused.
%


To validate the effectiveness of our \ourMthd{}, we conduct extensive experiments on a wide range of 3D objects of two popular benchmark datasets, \ie ShapeNet~\cite{chang2015shapenet} and Objaverse~\cite{deitke2023objaverse}, and an out-of-domain dataset, Google Scanned Objects~\cite{downs2022google}.
Based on the autoregressive manner and our well-designed strategies,\ourMthd{} can create high-quality 3D assets, as shown in \figref{fig:intro}.
The quantitative and qualitative results also demonstrate that our \ourMthd{} can significantly outperform recent cutting-edge 3D generation methods.

Our contributions can be summarized as follows:
\begin{itemize}




\item We present \ourMthd{}, a novel autoregressive pipeline composed of a 3D VAE and a 3D GPT for conditional 3D object generation.
To our knowledge, this is the first attempt to quantize the 3D objects with triplane representations and generate high-quality assets part by part.

\item We introduce feature deformation and attention mechanism designs to capture fine-grained geometry details.

\item We propose a 3D position encoding strategy, \ie \nameofposE{}, to preserve the spatial information as much as possible. 


\end{itemize}

\section{Related Work}
\label{sec:relatedwork}

\subsection{3D Generation}

%
Early 3D generation approaches mainly focus on the generation of different forms of 3D models, \eg point clouds~\cite{wu2016learning,ji2024jm3d,zhang24FastPCI}, meshes~\cite{chen2019learning,xu2023survey,zhu_2024_gsror,zhu_2025_dsdf}, and volumes~\cite{chan2022efficient,wang2023nice,sun2024recent} in a text/image-conditioned or unconditional manner.
Limited by the categories and quantity of 3D objects used for training, these methods often do not generalize well.
With the remarkable progress achieved by diffusion models in 2D image generation, a large number of researchers have explored migrating the pre-trained 2D priors to 3D generation.
Pioneer works~\cite{poole2022dreamfusion,wang2023score,chen2023fantasia3d,zhang2024temo} feed the rendered views to a 2D pre-trained model and perform per-shape optimization for knowledge distillation. 
Despite ushering in a new era, these methods suffer from a series of serious issues, \eg time-consuming and multi-face.
Another line of methods like Zero123/++~\cite{liu2023zero,shi2023zero123++}, One2345~\cite{liu2024one}, and AR123~\cite{zhang2025ar} use the rendered views of 3D objects to finetune the pre-trained diffusion models for new-view or multi-view generation.
To improve the consistency between the generated multiple views, Consistent123~\cite{lin2023consistent123} and Cascade-Zero123~\cite{chen2023cascade} introduce extra priors, \eg boundary, and redundant views. 
SyncDreamer~\cite{liu2023syncdreamer} proposes to correlate the corresponding features across different views by building a 3D-aware attention mechanism. 
These methods can be followed by a sparse-view reconstruction model, \eg NeRFs~\cite{mildenhall2021nerf,wang2021neus,muller2022instant}, LRMs~\cite{hong2023lrm,openlrm,xu2024instantmesh} and LGM~\cite{tang2024lgm}, to generate 3D objects.
However, the indirect generation fashion of these methods may lead to detail loss or reconstruction failures due to their heavy reliance on the fidelity of multi-view images.

More recently, a surge of works~\cite{zhang20233dshape2vecset,zhao2024michelangelo,wu2024direct3d,chen20253dtopia} directly synthesize 3D objects via a 3D diffusion model, in which the 3D shapes are fed to a pre-trained 3D VAE for continuous latent features.
Unlike these methods, our \ourMthd{} eschews the diffusion scheme and creates 3D objects by autoregressively generating discrete geometric parts.

\begin{figure*}[t]
  \centering
  \footnotesize
   \includegraphics[width=.98\linewidth]{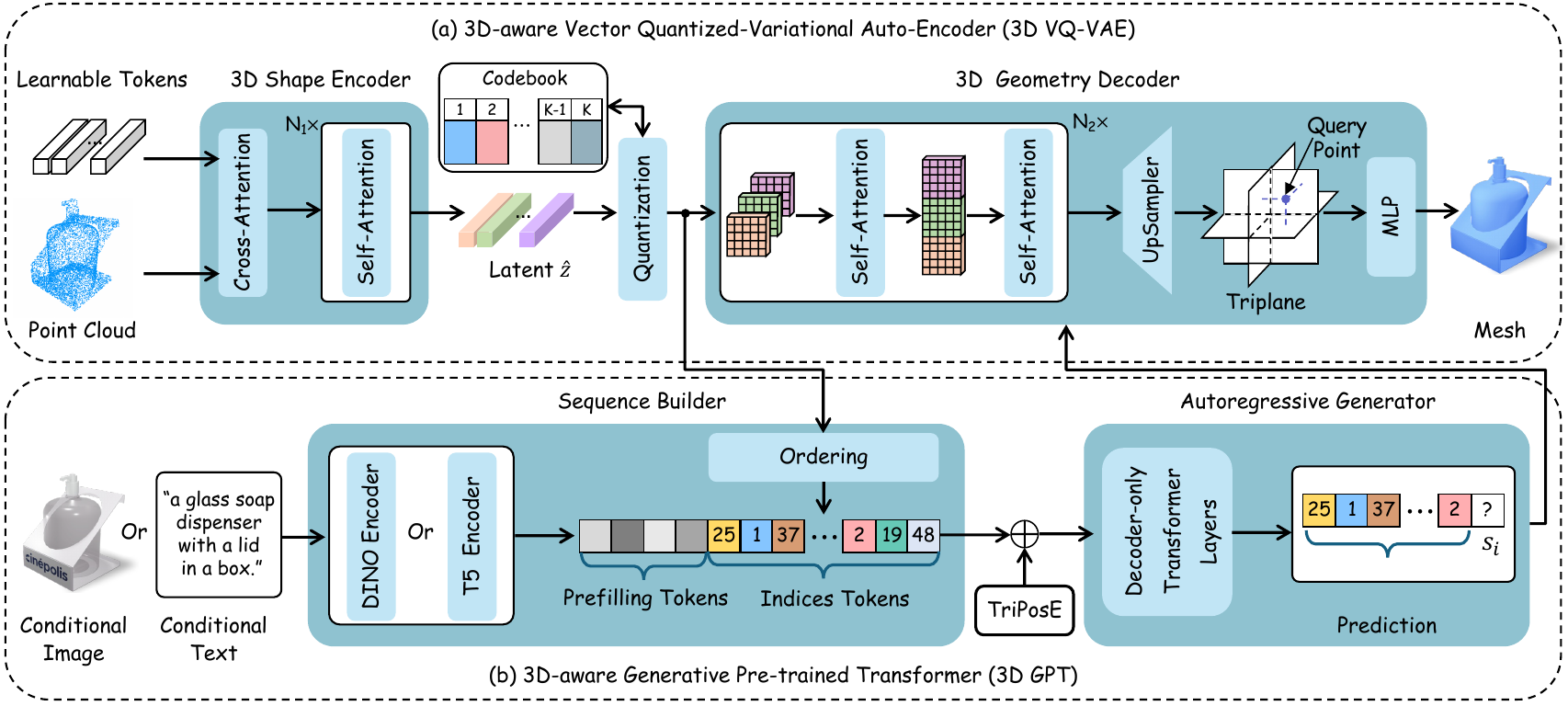}
    \vspace{-3pt}
   \caption{Overall architecture of the proposed \ourMthd{} framework. 
   (a) 3D VQ-VAE first encodes the point cloud uniformly sampled from 3D meshes into a set of learnable tokens in the triplane latent space. 
   Then, these continuous triplane features are quantized as discrete embeddings from a trainable codebook.
   Next, these quantized representations are deformed twice, along with two self-attention modules in several attention layers to achieve feature enhancement in each plane and information interaction among the three planes.
   Subsequently, the triplane features are upsampled to a higher resolution for fine-grained geometry details.
   Finally, the query point features sampled from this triplane are fed to an MLP network for their occupancy predictions.
   (b) 3D GPT first organizes the triplane indices from the pre-trained codebook of 3D VQ-VAE into a sequence, in which the indices within each plane are placed in a raster scan order and the indices at the same positions of the three planes in an adjacent order.
   Then, the prompt features are employed as the prefilling token embedding of the sequence for conditional 3D object generation.
   Next, this sequence is modeled by multiple decoder-only transformer layers via next-part prediction.
   By querying the codebook, the predicted index sequence can be transformed into triplane features to synthesize 3D objects.
   }
   \label{fig:overall}
   \vspace{-7pt}
\end{figure*}

\subsection{VAE \& VQ-VAE}
Variational Autoencoder (VAE)~\cite{kingma2013auto}
%
is usually utilized to map high-dimensional input information to continuous probabilistic latent representations and has a far-reaching impact in the field of generative modeling. 
Thanks to this work, recent diffusion models~\cite{rombach2022high,saharia2022photorealistic,chen2024region,tan2025auto} can be trained on limited computational resources while retaining their quality and flexibility.
Recent 3D generation approaches like Clay~\cite{zhang2024clay}, Direct3D~\cite{wu2024direct3d}, and LN3Diff~\cite{lan2025ln3diff} also explore to construct 3D-aware diffusion models based on this technology.
Vector Quantised-Variational AutoEncoder (VQ-VAE)~\cite{van2017neural} is a variant of VAE.
It introduces the codebook mechanism to quantize the continuous latent representations into discrete components, achieving promising performance on many generative tasks, \eg text-to-image generation~\cite{ramesh2021zero}, music generation~\cite{dieleman2018challenge,dhariwal2020jukebox}, and speech gesture generation~\cite{ao2022rhythmic}.
VQ-GAN~\cite{esser2021taming} proposes to promote VQ-VAE by incorporating adversarial loss during the training process.
In this paper, we build a 3D-aware VQ-VAE to quantize the triplane features of 3D shapes and acquire discrete geometric parts for autoregressive 3D generation.
%

\subsection{GPT}
Generative Pre-trained Transformer (GPT)~\cite{radford2018improving} originates in the field of natural language processing (NLP).
%
Based on the decoder-only transformer architecture, it autoregressively generates text sequences according to the next-token-prediction paradigm. 
This series of works~\cite{zhang2021rstnet,achiam2023gpt,bai2023qwen,chen2024internvl,li2024promptkd} with groundbreaking reasoning ability and incredible scalability continue to emerge, revolutionizing language generation.
Inspired by these achievements, many researchers have attempted to transfer this scheme to image generation.
For example, Parti~\cite{yu2022scaling} proposes a pathways autoregressive text-to-image model, which regards image generation as sequence-to-sequence modeling for high-fidelity images.
%
LlamaGen~\cite{sun2024autoregressive} verifies that vanilla autoregressive models, \eg Llama~\cite{touvron2023llama}, can achieve state-of-the-art image generation performance without inductive biases on visual signals if scaling properly.
Emu3~\cite{wang2024emu3} achieves excellent performance in both generation and perception tasks by tokenizing images, text, and videos into a discrete space.
Besides, AutoSDF~\cite{mittal2022autosdf} learns a `non-sequential' autoregressive shape prior for 3D completion, reconstruction, and generation.
More recently, several works like MeshGPT~\cite{siddiqui2024meshgpt}, MeshAnything~\cite{chen2024meshanything}, and MeshXL~\cite{chen2024meshxl} also explore to autoregressively generate the faces of 3D meshes with GPT.

In contrast to these methods, our \ourMthd{} quantizes the triplane representations of 3D shapes into discrete geometric parts and uses the GPT model to generate 3D objects in a next-part prediction manner.

\section{Methodology}


\figref{fig:overall} illustrates the overall architecture of our \ourMthd{} framework.
Our goal is to migrate the promising learning and multimodal unification capabilities of GPT~\cite{achiam2023gpt} to conditional 3D object generation.
However. existing methods~\cite{siddiqui2024meshgpt,chen2024meshanything,chen2024meshxl} quantizing the mesh faces suffer from excessively long sequences, limiting their applications for high-quality 3D assets.
In this work, we propose to represent the 3D shape information of meshes with triplane latent representations whose feature maps are associated with three-axis planes, \ie XY, YZ, and XZ.
These triplane features can be quantized by a trainable codebook, resulting in a fixed length sequence, regardless of the number of faces.

\subsection{3D VQ-VAE} \label{subsec:3dvqvae}

To quantize the triplane representation of 3D shapes into discrete embeddings and synthesize 3D objects, we develop a 3D VQ-VAE, including a 3D shape encoder, a quantizer, and a 3D geometry decoder.

\myPara{3D shape encoder} is designed to acquire compact and robust latent representations of 3D objects for fine-grained geometry information.
Given a 3D object, we first uniformly sample high-resolution point clouds from its surface.
To enhance the expressive power of the point clouds, the corresponding normals are also included in the point cloud representation that is denoted as $\mathrm{P} \in \mathbb{R}^{B \times N_{p} \times (3+3)}$, where $B$ is the batch size, and $N_{p}$ is the number of points.
Then, we apply the Fourier positional ~\cite{tancik2020fourier} encoding to the point cloud representation to capture high-frequency details.

Inspired by previous works for point cloud understanding~\cite{jaegle2021perceiver,zhang20233dshape2vecset}, we employ a transformer-based architecture consisting of a cross-attention layer and $N_{1}$ self-attention layers to extract the latent features of the 3D point cloud.
More precisely, the point cloud information is injected into a series of learnable query tokens that is denoted as $\mathrm{e} \in \mathbb{R}^{B \times (3 \times h \times w) \times d_{e}}$, where $h$ and $w$ are the height and width of the triplane feature maps respectively, and $d$ represents the channel number of these learnable tokens, via the cross-attention layers.
After that, the representational ability of these tokens is enhanced by the following self-attention layers, resulting in triplane latent representations, \ie $\hat{\mathrm{z}} \in \mathbb{R}^{B \times (3 \times h \times w) \times d_{z}}$.

\myPara{Quantizer} is introduced to represent the continuous triplane features with the embeddings $z_{q}$ from a learnable, discrete codebook $\mathcal{Z} = \{ z_{k} \}_{k=1}^{K} \subset \mathbb{R}^{d_{q}}$.
Specifically, we first use a linear layer to project the continuous features to the same channel number with the codebook embeddings, yielding features $\tilde{\mathrm{z}} \in \mathbb{R}^{B \times (3 \times h \times w) \times d_{q}}$.
Then, an element-wise quantization between each spatial code of these features and its closest codebook entry is performed as follows:
\begin{equation}
  z_{q} := \bigg ( \mathop \mathrm{arg~min}_{z_{k} \in \mathcal{Z}} || \tilde{\mathrm{z}}_{ij} - z_{k}|| \bigg ) \in \mathbb{R}^{B \times (3 \times h \times w) \times d_{q}}.
  \label{eq:vq}
\end{equation}

\myPara{3D geometry decoder} aims to reconstruct the 3D neural field in a high quality with the quantized discrete features, \ie $z_{q}$, as input.
Inspired by recent works in video generation~\cite{guo2023animatediff,hu2024animate} and avatar generation~\cite{wang2023rodin}, we achieve plane information interaction by feeding $z_{q}$ to $N_{2}$ attention layers consisting of two feature deformations and self-attention operations.
In particular, the plane axis is ignored by being reshaped into the batch axis, allowing the first self-attention to process each plane independently.
The plane axis is recovered and the features of the three planes are concatenated along the height dimension, enabling the second self-attention to model the information interaction across different planes.
We also upsample the triplane features to a higher resolution following the operation in Direct3D~\cite{wu2024direct3d}.
Given a set of query points consisting of voxel points and near-surface points in the 3D field, their features can be sampled from the yielding triplane via bilinear interpolation.
The query point features are transformed to their occupancy values via a Multi-Layer Perceptron (MLP).


\begin{figure}[t]
  \centering
  \footnotesize
   \includegraphics[width=0.98\linewidth]{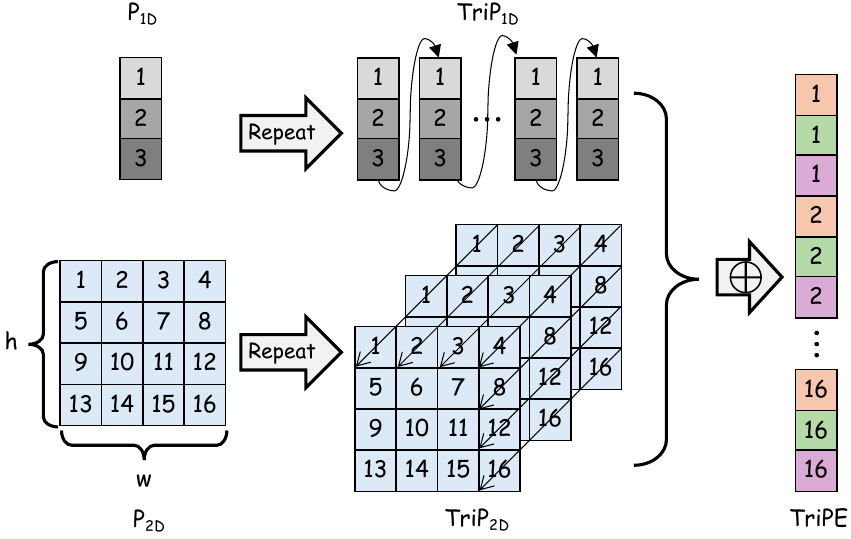}
   \vspace{-3pt}
   \caption{
        Diagrammatic details of our \nameofposE{} designed for the positional encoding of triplane sequence.
We represent the positional information with numbers and simplify the number of tokens in 2D encoding for ease of presentation. }
   \label{fig:tripose}
   \vspace{-8pt}
\end{figure}

\subsection{3D GPT} \label{subsec:3dgpt}

To model the constituents of a 3D object in an autoregressive manner, we propose a 3D GPT, which consists of three components, \ie a sequence builder, a \nameofposE{} position encoding, and an autoregressive generator.

\myPara{Sequence builder.}
Based on the pre-trained 3D VQ-VAE, the 3D shapes are encoded as discrete triplane features, which can be represented with their indices in the codebook.
Subsequently, these indices are transformed into a sequence according to some ordering rules.
Considering the triplane representation is composed of three correlated feature maps, we organize the indices within each plane in a raster scan order and the indices at the same positions of the three planes in the adjacent orders.
To achieve conditional 3D generation, the prompts are encoded as the prefilling token embedding of the sequence.

\myPara{\nameofposE{}} is a 3D position encoding strategy tailored for the triplane index sequence.
As shown in \figref{fig:tripose}, it is a fusion of 2D position encoding and 1D position encoding based on the  Rotary Position Embedding (RoPE)~\cite{su2024roformer}.
We denote the RoPE for a 2D feature map with height $h$ and width $w$ as $\mathrm{P_{2D}} \in \mathbb{R}^{h \cdot w}$, and the RoPE for a 1D sequence with 3 tokens as $\mathrm{P_{1D}} \in \mathbb{R}^{3}$.
Note that the channel dimension is removed for ease of description.
To preserve the 2D spatial information in the axis-aligned feature planes for the triplane index sequence, we repeat the unit element of $\mathrm{P_{2D}}$ three times and place the two newly emerged elements adjacent to their original element.
We denote this 2D position encoding for triplane indices as $\mathrm{TriP_{2D}} \in \mathbb{R}^{3 \cdot h \cdot w}$.
Meanwhile, we repeat the three elements in $\mathrm{P_{1D}}$ for $h \times w$ times to highlight the difference of the three feature maps, yielding a 1D position encoding for triplane indices denoted as $\mathrm{TriP_{1D}} \in \mathbb{R}^{3 \cdot h \cdot w}$.
Finally, we calculate the \nameofposE{} by performing element-wise addition of $\mathrm{TriP_{2D}}$ and $\mathrm{TriP_{1D}}$.

\myPara{Autoregressive generator.}
After tokenizing the triplane indices into a sequence $s \in \{ 0, ..., K \mathrm{-} 1, K\}^{3 \cdot h \cdot w}$, along with the prompts $c$ and custom positional encoding \nameofposE{}, 3D object generation can be formulated as an autoregressive next-index prediction.
To be specific, the decoder-transformer layers learn to predict the distribution of possible next indices, which can be written as follows:
\begin{equation}
  p_{\theta}(s|c) = \prod \limits_{t} p_{\theta}(s_{t}|s_{<t}, c),
  \label{eq:ar_pred}
\end{equation}
where $t$ is the time step in the generation process, $c$ is a conditional image or text embedding, and $p_{\theta}$ denotes the decoder-transformer layers with parameters $\theta$.

\subsection{Optimization Details} \label{subsec:optim}
Corresponding to the overall architecture in \figref{fig:overall}, the optimization process of our \ourMthd{} framework can also be divided into two stages, \ie 3D VQ-VAE optimization and 3D GPT optimization.

To train our 3D VQ-VAE in an end-to-end manner, we employ the Binary Cross-Entropy (BCE) loss as the optimization objective for reconstructing 3D objects.
This process can be formalized as follows:
\begin{equation}
  \mathcal{L}_{rec} = \mathbb{E}_{x\in\mathbb{R}^{3}} \bigg [ \mathrm{BCE} \Big(\hat{\mathcal{O}}(x), \mathcal{O}(x)\Big) \bigg ],
  \label{eq:recloss}
\end{equation}
%
where $\hat{\mathcal{O}}(\cdot)$, and $\mathcal{O}(\cdot)$ are the predicted occupancy value and ground-truth occupancy value of the query point.
For the codebook learning of quantizer, the training loss can be formulated to minimize the difference between the original features and the quantified features: 
\begin{equation}
  \mathcal{L}_{cb} = || sg(\tilde{\mathrm{z}}) - z_{q} ||_{2}^{2} + \beta || \tilde{\mathrm{z}} - sg[z_{q}]||,
  \label{eq:codebookloss}
\end{equation}
where $sg[\cdot]$ denotes the stop-gradient operation~\cite{bengio2013estimating}, and $\beta$ is a weight hyperparameter to balance the two-part losses, which is set as $\beta=0.25$ by default.
%
Finally, our 3D VQ-VAE is optimized by minimizing:
%
\begin{equation}
  \mathcal{L}_{3dvqvae} = \lambda_{rec} \mathcal{L}_{rec} + \lambda_{cb} \mathcal{L}_{cb},
  \label{eq:vqvaeloss}
\end{equation}
where $\lambda_{rec}$ and $\lambda_{cb}$ are the weights of reconstruction optimization and codebook optimization, respectively.

The optimization objective of our 3D GPT is to maximize the log-likelihood of the triplane index sequence.
As a result, its training loss can be expressed as follows:
\begin{equation}
  \mathcal{L}_{3dgpt} = - \sum_{t=1}^{3 \cdot h \cdot w} \log \big( p_{\theta}(s_{t}|s_{<t}, c) \big).
  \label{eq:arloss}
\end{equation}

\section{Experiments}

\subsection{Experiment Setup}\label{subsec:setting}
\myPara{Datasets.}
To examine the effectiveness of our \ourMthd{}, we conduct experiments on two benchmark 3D datasets, \ie ShapeNet~\cite{chang2015shapenet}, Objaverse~\cite{deitke2023objaverse}, and an out-of-domain dataset, \ie  Google Scanned Object(GSO)~\cite{downs2022google}.
Specifically, the ShapeNet dataset provides 52,472 manufactured meshes covering 55 categories.
We follow the splits from 3DILG~\cite{zhang20223dilg}, where 48,597 samples are used for training, 1,283 for validation, and 2,592 for testing.
Inspired by previous works on data filtering~\cite{li2023instant3d,zhang2024clay}, we score the rendered normal maps of 800,000 meshes in the Objaverse dataset and obtain about 100,000 geometry objects.
Moreover, 1,000 samples are randomly selected for performance evaluation, and the remaining ones are employed for model training.
In addition, the GSO dataset contains about 1000 real-world 3D scans, which are utilized to further validate the generalization of our method.
For each 3D asset in these two datasets, we adopt the rendered images and textual descriptions from ULIP~\cite{xue2023ulip} to build the prompt system.
We uniformly select 4 images from all the rendered images, whose top 1 captions are utilized as the text prompts.

\myPara{Metrics.}
We evaluate the performance of the methods from two aspects: 2D visual quality and 3D geometric quality.
For the 2D visual evaluation, we compare the novel views rendered from the synthesized 3D mesh with the ground truth views based on a set of common metrics, including Peak Signal-to-Noise Ratio (PSNR), Perceptual Loss (LPIPS)~\citep{zhang2018unreasonable}, Structural Similarity (SSIM)~\citep{wang2004image}, and CLIP~\cite{radford2021learning} score.
For the 3D geometric evaluation, we compare the point clusters that are randomly sampled from the generated meshes and the ground-truth meshes.
Following the protocol in previous works~\cite{xu2024instantmesh,liu2024meshformer}, we employ the chamfer distance value and the F-Score with a threshold of 0.02 as the primary evaluation metrics.

\begin{figure*}[t]
  \centering
  \footnotesize
   \includegraphics[width=\linewidth]{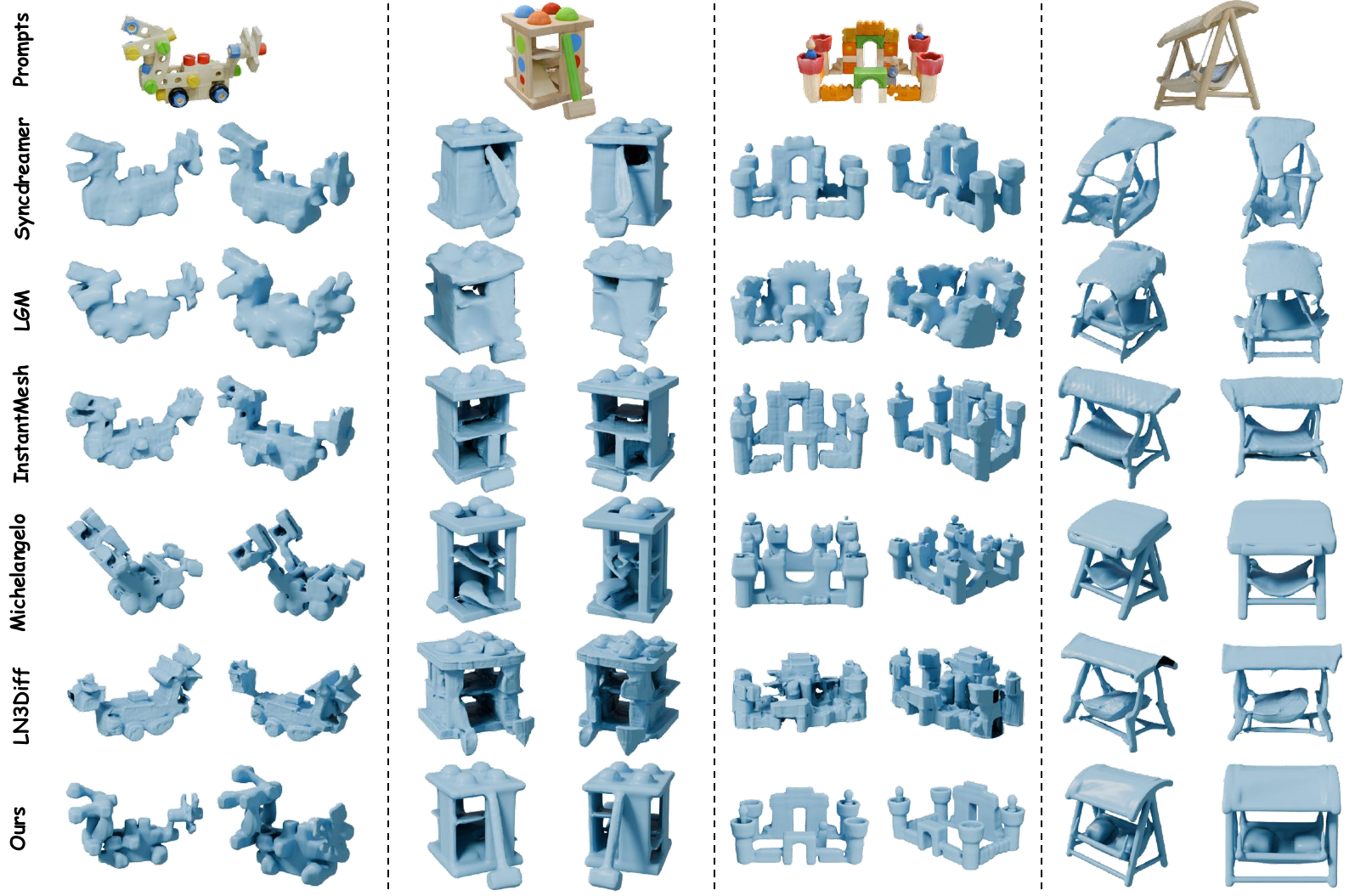}
    \vspace{-15pt}
   \caption{
   Visual comparisons of the 3D meshes generated by our \ourMthd{} and recent multiview-based models, \ie Syncdreamer, LGM and InstantMesh, and  3D native approaches,\ie Michelangelo and LN3DIff,  for image-to-3D object generation.
   Given the same input images from the GSO dataset, our \ourMthd{} can produce 3D assets with superior geometric details against other baselines.
   }
   \vspace{-5pt}
   \label{fig:ito3d}
\end{figure*}

\myPara{Implementation details.}
In our 3D VQ-VAE, the number of point clouds input to the 3D shape encoder, \ie $N_{p}$, is 81,920.
The number of the self-attention layer in the 3D shape encoder, \ie $N_{1}$ is 8.
The height and width of the triplane feature maps, \ie $h$ and $w$, are both set to 32. 
The channel numbers of the learnable tokens, and the triplane features, \ie $d_{e}$ and $d_{z}$, are set to 768 and 16. 
The size and channel number of the codebook, \ie $K$ and $d_{q}$, are set to 16,384 and 8, respectively.
Moreover, the number of the attention layers in the 3D geometry decoder is set as $N_{2} \mathrm{=} 6$. 
The triplane features are upsampled to a resolution of $256 \mathrm{\times} 256$, in which 20,480 uniformly sampled voxel points and 20,480  near-surface points are employed as query points for occupancy supervision.
The hyperparameters in \equref{eq:vqvaeloss} are set as $\lambda_{rec} \mathrm{=} 1$ and $\lambda_{cb} \mathrm{=} 0.1$.
We employ the CosineAnnealing scheduler~\cite{loshchilov2016sgdr}, where the learning rate is initialized to $1e\mathrm{-}4$ and gradually decays over time.
We adopt the AdamW optimizer~\cite{loshchilov2019decoupled} to train the 3D VQ-VAE with a total batch size of 128 for 100K steps on 8 NVIDIA A100 GPUs.

In our 3D GPT, we adopt the pre-trained DINO~\cite{zhang2022dino}(ViT-B16) and FLAN-T5 XL~\cite{chung2024scaling} to encode the conditional images and text prompts respectively.
As far as the decoder-only transformer, we follow the GPT-L setting of LLamaGen~\cite{sun2024autoregressive}, which consists of 24 transformer layers with a head number of 16 and a dimension of 1024.
We employ the AdamW optimizer with a learning rate of $1e \mathrm{-} 4$ and a total batch size of 80 to train our 3D GPT for 100K steps.
In addition, the classifier-free guidance (CFG)~\cite{ho2022classifier} scale of 7.5 is also introduced in the autoregressive inference to improve geometry quality and image/text-3D alignment.

\subsection{Qualitative Evaluation}\label{subsec:qual}


\myPara{Image-to-3D.}
We first provide visualization comparisons between our \ourMthd{} with recent \sArt models for image-to-3D generation, including three multiview-based methods like SyncDreamer~\cite{liu2023syncdreamer},  InstantMesh~\cite{xu2024instantmesh}, OpenLRM~\cite{openlrm}, and LGM~\cite{tang2024lgm} and
three 3D native  approaches like Shap-E~\cite{jun2023shap}, Michelangelo~\cite{zhao2024michelangelo} and LN3Diff~\cite{lan2025ln3diff}.
As shown in the garden-swing sample of \figref{fig:ito3d}, the methods based on mulit-view synthesis may produce 3D objects with discontinuous geometry parts or even incorrect objects, which are attributed to inconsistencies in their views used for reconstruction.
Meanwhile, the LEGO sample shows that 3D diffusion methods are prone to generating noisy 3D objects, possibly due to anomalies in the conditional denoising process.
In contrast to these approaches, our \ourMthd{} model, equipped with the powerful learning capabilities of GPT scheme, can synthesize high-quality 3D objects with superior geometric details.

\begin{figure*}[t]
  \centering
  \footnotesize
   \includegraphics[width=.98\linewidth]{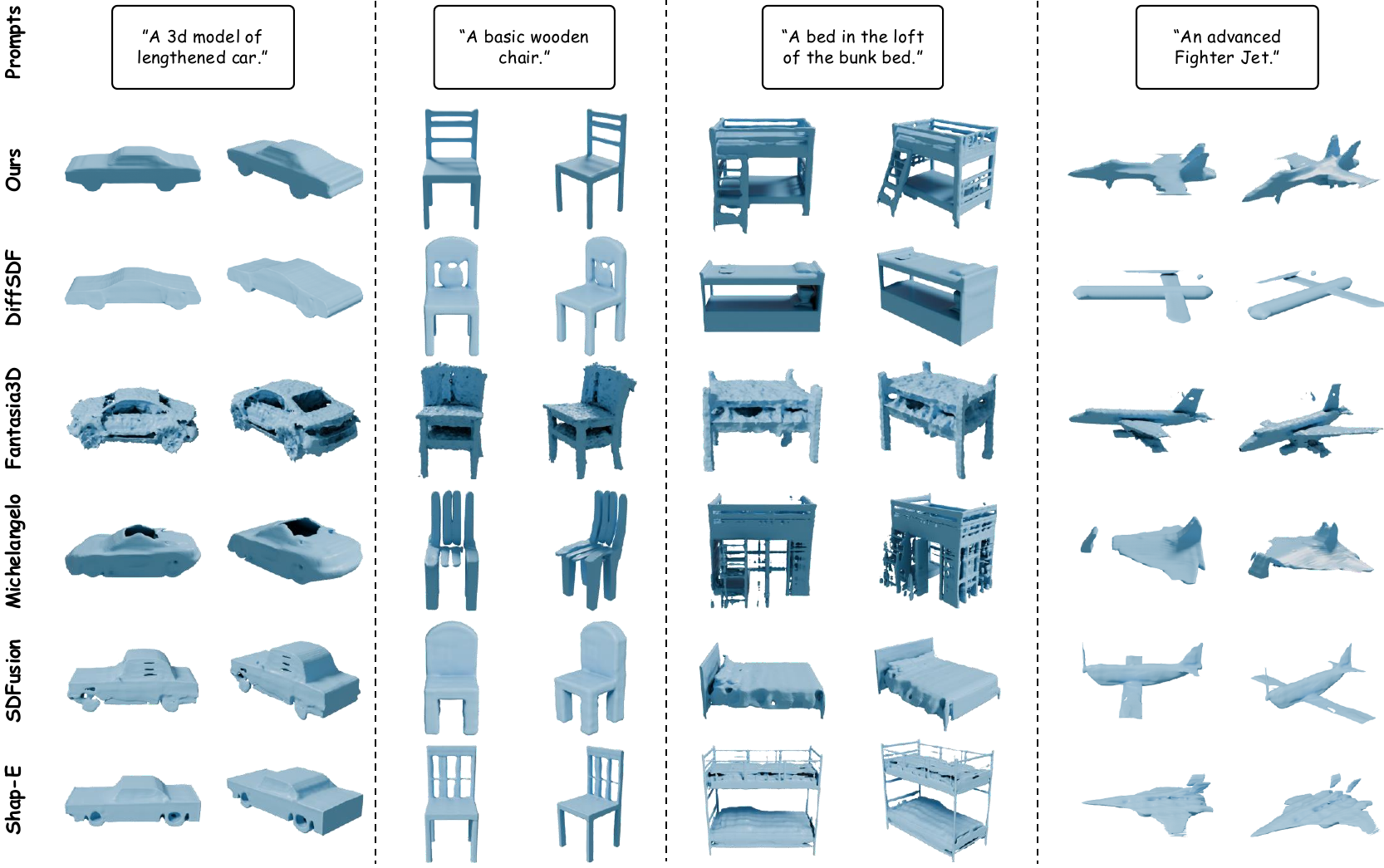}
    \vspace{-5pt}
   \caption{Qualitative comparisons of our \ourMthd{} with recent cutting-edge methods for the text-to-3D object generation task.
   The 3D mesh assets created by our \ourMthd{} are semantically more faithful to the given textual prompts compared to previous approaches.
   }
   \label{fig:tto3d}
\end{figure*}
\begin{table*}[tp]
\setlength\tabcolsep{5pt}
\centering
\footnotesize
\caption{
Quantitative comparisons of our \ourMthd{} model with recent \sArt image-to-3D generation methods on 2D visual quality and 3D geometric quality.
`$\uparrow$': the higher the value, the better the performance, `$\downarrow$': the lower the better.
} 
\vspace{-5pt}
\begin{tabular}{l|ccccc|c}
\toprule
    Methods & Shap-E~\cite{jun2023shap} & SyncDreamer~\cite{liu2023syncdreamer} & Michelangelo~\cite{zhao2024michelangelo} & InstantMesh~\cite{xu2024instantmesh} & LGM~\cite{tang2024lgm} &    \ourMthd{} (Ours)\\
\midrule 
PSNR~$\uparrow$  &  10.991 &  11.269 & 11.928  & 11.560  & 11.363 & 13.626 \\
SSIM~$\uparrow$  &  0.702 &  0.706 &  0.734 & 0.721 & 0.714 &  0.763 \\
Clip-Score~$\uparrow$  & 0.834 & 0.837 & 0.864  & 0.847   & 0.841 & 0.868 \\
LPIPS~$\downarrow$  & 0.325 &  0.320 &  0.278 & 0.303& 0.317 & 0.216 \\
\midrule 
Chamfer Distance~$\downarrow$  & 0.156 & 0.158  &  0.117 &  0.137 &  0.149 & 0.066 \\
F-Score~$\uparrow$  &  0.163 & 0.178  &  0.226 & 0.179  &  0.172 & 0.303\\
\bottomrule
\end{tabular}
\label{tab:quan}
\end{table*}

\myPara{Text-to-3D.}
We compare our \ourMthd{} with other cutting-edge text-to-3D approaches, including Diffusion-sdf~\cite{li2023diffusion}, SDFusion~\cite{cheng2023sdfusion}, Shap-E~\cite{jun2023shap}, Fantasia3D~\cite{chen2023fantasia3d}, and Michelangelo~\cite{zhao2024michelangelo}.
As shown in \figref{fig:tto3d}, these baseline methods are susceptible to failure in various cases, either generating poor quality 3D objects, like the result of Michelangelo for ``An advanced
fighter Jet'', or not matching the given textual descriptions, like the result of Fantasia3D for ``a bed in the loft of the bunk bed''.
Different from them, our \ourMthd{} can create significantly more plausible 3D assets through GPT autoregression driven by text prefilling embeddings.

\subsection{Quantitative Evaluation}\label{subsec:quan}

In this subsection, we use the mixed evaluation set of ShapeNet and Objaverse to quantitatively measure the performance of our \ourMthd{} model against recent 3D object generation methods in terms of 2D visual quality and 3D geometric quality.
Considering the diversity of 3D objects generated from text descriptions, we conduct experiments on image-to-3D tasks to ensure an accurate comparison, as shown in \tabref{tab:quan},.
In particular, all candidate approaches employ the same images as their conditional input to generate 3D meshes.
For the 2D evaluation, we render 20 views with $224 \times 224$ resolution for each mesh and compare the yielding normal maps with the ones of corresponding ground-truth views.
For the 3D evaluation, we compare the generated meshes with the ground truth meshes by uniformly sampling 16K point clouds from their surfaces in an aligned cube coordinate system of $[-1, 1]^{3}$.
Our \ourMthd{} model surpasses
other cutting-edge 3D object generation methods across all 2D and 3D metrics by a large margin.
These experimental results further demonstrate the superiority of our \ourMthd{} over existing methods.

\begin{table*}[tp]
    \centering
    \footnotesize
    
    \begin{minipage}{.29\textwidth}
        \centering
        \setlength\tabcolsep{2pt}
        \caption{Ablation studies on the reconstruction capability of the 3D VQ-VAE.} 
        \vspace{-7pt}
        \begin{tabular}{lcc} \toprule
            & 3D VAE & 3D VQ-VAE \\  
            \midrule 
            Chamfer Distance~$\downarrow$  &  0.018 & 0.016  \\
            \midrule 
            F-Score~$\uparrow$  & 0.811 &  0.822 \\
            \bottomrule
            \end{tabular}
            \vspace{-4pt}
            \label{tab:vqvae}
    \end{minipage}%
    \hfill
    \begin{minipage}{.28\textwidth}
        \centering
        \setlength\tabcolsep{4pt}
        \caption{Ablation studies on the PII design of the 3D-VQVAE.} 
        \vspace{-7pt}
        \begin{tabular}{lcc} \toprule
        & \emph{w/o} PII & \emph{w/} PII  \\  
        \midrule 
        Chamfer Distance~$\downarrow$ & 0.023  &  0.016   \\
        \midrule 
         F-Score~$\uparrow$ & 0.661 & 0.822  \\
        \bottomrule
        \end{tabular}
        \vspace{-3pt}
        \label{tab:abl_pii}
    \end{minipage}%
    \hfill
    \begin{minipage}{.4\textwidth}
        \centering
        \setlength\tabcolsep{3pt}
        \caption{Ablation studies on triplane size (\ie sequence length) of the 3D GPT.} 
        \vspace{-7pt}
        \begin{tabular}{lccc} \toprule
        Triplane Size & $3 \!\times\! 16 \!\times\! 16$ & $3 \!\times\! 32 \!\times\! 32$ & $3 \!\times\! 48 \!\times\! 48$ \\    
        \midrule 
        Chamfer Distance~$\downarrow$  & 0.157  &  0.066 & 0.062 \\
        \midrule 
        Inference Time~$\downarrow$  &  17.7~s &  67.6~s & 143.9~s\\
        \bottomrule
        \end{tabular}
        \vspace{-3.5pt}
        \label{tab:abl_triplanesize}
    \end{minipage}%
\end{table*}


\subsection{Ablation Studies} \label{subsec:abl}

\myPara{Evaluation on 3D VQ-VAE.}
We evaluate the 3D reconstruction capability of our 3D VQ-VAE, which serves as a foundation for generating high-quality 3D assets.
For reference, we provide the performance of the VAE counterpart from which our 3D VQ-VAE is derived.
To obtain a well-performing tokenizer for autoregressive generation, we adjust the training strategies until the performance of our 3D VQ-VAE is on par with or better than that of the VAE counterpart.
The experiment results shown in \tabref{tab:vqvae} demonstrate the reconstruction effectiveness of our 3D VQ-VAE.

\myPara{Plane information interaction in 3D VQ-VAE.}
We analyze the importance of plane information interaction (PII) achieved by the feature deformation and attention mechanism in our 3D VQ-VAE.
As shown in \tabref{tab:abl_pii}, the variant without PII design achieves $0.661$ f-score in 3D reconstruction.
With the incorporation of our PII designs, this score significantly increased to $0.822$.
These experiments indicate that our PII designs contribute to the decoding process from the latent features to 3D objects.

\begin{figure}[t]
  \centering
  \footnotesize
   \includegraphics[width=0.94\linewidth]{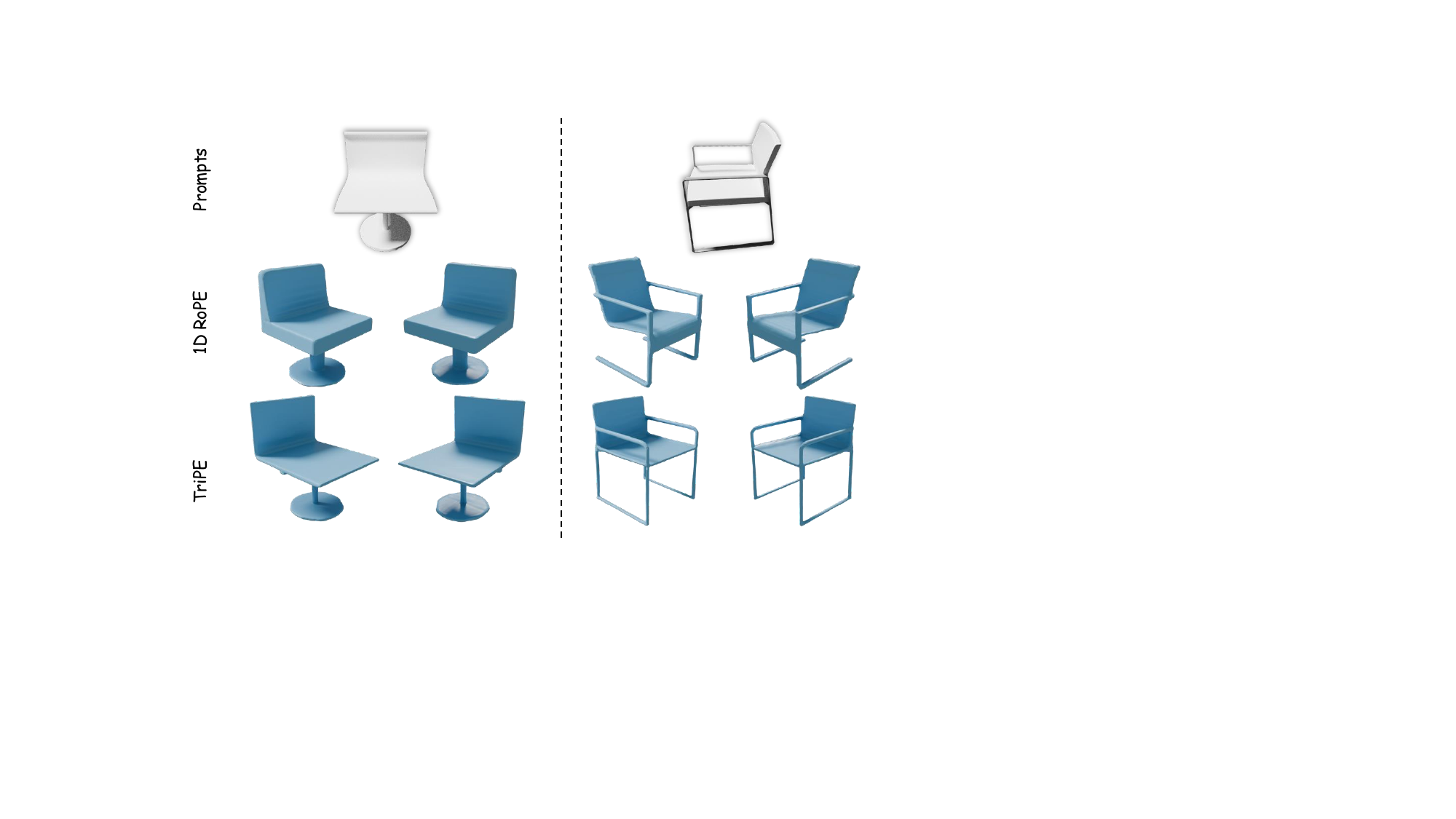}
   \vspace{-7pt}
   \caption{Ablation experiments on the effectiveness of our \nameofposE{}.}
   \label{fig:abl_pos}
   \vspace{-8pt}
\end{figure}

\myPara{\nameofposE{} positional encoding in 3D GPT.}
We investigate two positional encoding strategies for the sequence modeling in 3DGPT.
The first strategy is the 1D Rotary Position Embedding (RoPE)~\cite{su2024roformer} matching the sequence length, and the other is the proposed \nameofposE{}.
As can be observed in \figref{fig:abl_pos}, the 1D RoPE is prone to lose important geometric details of the objects in the input images.
In contrast, our well-designed \nameofposE{} can preserve as much 3D spatial information as possible, thereby generating 3D objects that are geometrically more faithful to the prompts.

\begin{figure}[t]
  \centering
  \footnotesize
   \includegraphics[width=0.94\linewidth]{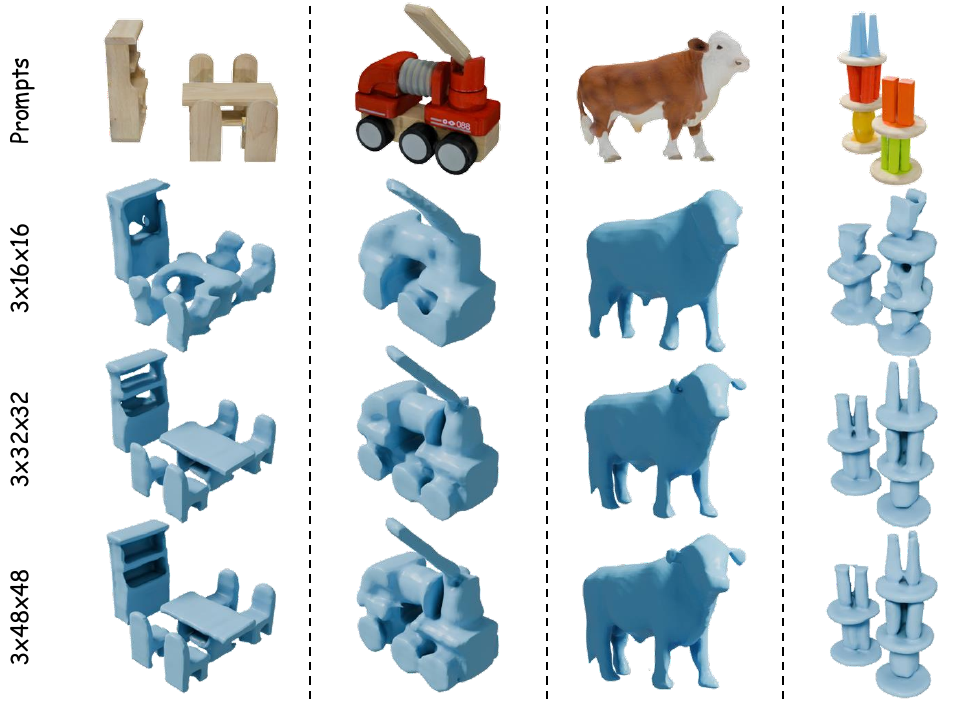}
   \vspace{-7pt}
   \caption{
        Visual comparisons of our \ourMthd{} variants with different triplane sizes or sequence lengths. }
   \label{fig:triplanesize}
   \vspace{-9pt}
\end{figure}

\myPara{Triplane Size.}
We ablate the triplane size with 3$\times$16$\times$16, 3$\times$32$\times$32, and 3$\times$48$\times$48 to explore the impact of sequence length on the performance and efficiency.
Note that the 3$\times$ 32$\times$32 widely adopted in 3D reconstruction is our default setting.
As shown in \tabref{tab:abl_triplanesize}, it is the best setting for efficiency and performance trade-offs, which is also demonstrated by the visual comparisons in \figref{fig:triplanesize}

\section{Conclusions \& Future Work}

In this paper, we propose a novel framework named \ourMthd{} to generate high-quality 3D assets via the ``next-token prediction'' paradigm derived from multimodal large language models.
To achieve this, we first develop a 3D VQ-VAE, in which the 3D shapes are encoded into the triplane latent space and quantized as discrete embeddings from a trainable codebook.
Equipped with the well-designed feature deformation and attention mechanism, these quantized features are utilized to reconstruct fine-grained geometries in a neural occupancy field.
Then, we adopt the codebook indices of these discrete representations to form the sequence for autoregressive modeling.
Driven by the prefilling prompt embeddings, our 3D GPT, along with the 3D spatial information from the proposed \nameofposE, can generate high-quality 3D objects part by part.
%

We believe it is promising to improve 3D object generation under the autoregressive setting.
There are a few foreseeable directions to explore:
1) Scaling law. We will collect more 3D data with prompts from other datasets, \eg Objaverse-XL~\cite{deitke2024objaverse}, 3D-FUTURE~\cite{fu20213d}, and scale the GPT model.
2) Sequence formulation. We will explore the sequence formulation manner of triplane indices for more efficient generative modeling.

\myPara{Acknowledgments.}
This research was partially funded by Shenzhen Science and Technology Program (No. JCYJ20240813114237048).

{
    \small
    \bibliographystyle{ieeenat_fullname}
    \bibliography{main}
}


\end{document}